\documentclass{article}
\usepackage[utf8]{inputenc}

\usepackage{amsmath}
\usepackage{amssymb}
\usepackage{amsfonts}

\usepackage{url}

\usepackage{listings}

\usepackage{graphicx}

\usepackage{hyperref}
\usepackage{cleveref}

\title{Efficient Sampled Softmax Loss in Tensorflow}
\author{Maciej Skorski}

\begin{document}

\maketitle

\begin{abstract}
This short paper discusses an efficient implementation of \emph{sampled softmax loss} for Tensorflow. The speedup is achieved due to simplification of the graph for the forward and backward passes.
\end{abstract}

\section{Introduction}

The softmax function is used in prediction and classification tasks to map outputs of a network into probabilities. The corresponding formula reads
\begin{align*}
y_c = \frac{\exp(o_c)}{\sum_{c'}\exp(o_{c'})}
\end{align*}
where $c$ is the output class of interest, $o_{\cdot}$ are network outputs and the summation over $c'$ is taken over all possible classes. It is typically trained under cross-entropy loss. Unfortunately computing the loss is computationally expensive because of \emph{explicit normalization}. The factor in the denominator runs over all classes which may be quite large ($10^{5}$ and more for NLP problems). 

The solution is to approximate the loss function. One strategy, called \emph{sampled softmax}~\cite{jean2014using}, is to compute softmax over  a random subsample containing the target (true) class. 

In this note we present a more efficient implementation of the coupled \emph{sampled softmax}+\emph{cross entropy} loss for the leading machine learning framework TensorFlow~\cite{tensorflow2015-whitepaper}. The code is available online\footnote{See the GitHub repo \url{https://github.com/maciejskorski/ml_examples/blob/master/efficient_sampled_softmax.ipynb}}.

\section{Implementation}

We find that the existing implementation \texttt{tf.nn.sampled\_softmax\_loss} from TensorFlow~\cite{tf_softmax} produces a graph which is overly complicated. Simplifying this graph we obtain a considerable improvement. We also simplify and explicitly calculate the gradients of the composed loss function, instead of relying on auto-differentiation. Tests are provided for correctness of both: forward and backward passes.

\subsection{Implementation of Forward Pass}

\lstset{basicstyle=\small}
\begin{lstlisting}
with tf.variable_scope('sampled_softmax'):

    with tf.variable_scope('target_embeddings'):
      samples_embed = tf.gather(target_embed_kernel,samples) # (N_SAMPLED,N_EMBED)
      labels_embed = tf.gather(target_embed_kernel,labels,axis=0) # (N_BATCH,N_EMBED)
    
    with tf.variable_scope('labels_logits'):
      labels_logits = tf.matmul(tf.expand_dims(labels_embed,1),tf.expand_dims(inputs_embed,-1)) # (N_BATCH,1,1)
      labels_logits = tf.squeeze(labels_logits,-1) # N_BATCH,1
      labels_logits = labels_logits-tf.log(labels_prior) # add prior-correction
    
    with tf.variable_scope('sampl_logits'):
      samples_logits = tf.matmul(inputs_embed,samples_embed,transpose_b=True) # (N_BATCH,N_SAMPLED)
      samples_logits = samples_logits-tf.log(samples_prior) # add prior-correction
      
    with tf.variable_scope('sampl_loss'):
      candidate_logits = tf.concat([samples_logits,labels_logits],axis=-1) # (N_BATCH,N_SAMPLED+1)
      Z = tf.reduce_logsumexp(candidate_logits,axis=-1,keepdims=True) # (N_BATCH,1)
      loss = tf.reduce_mean(-labels_logits+Z)
    
    return loss
\end{lstlisting}

\subsection{Implementation of Backward Pass}

\lstset{basicstyle=\small}
\begin{lstlisting}
  with tf.variable_scope('backprop'):
    # note: gradients computed as sparse slices (tf.IndexedSlice), indices may duplicate
    grad_input_embed_shape,grad_target_embed_shape = gradient_shapes
    batch_len = tf.cast(tf.shape(labels)[0],tf.float32)
    samples_pred = tf.exp(samples_logits-Z)/batch_len # (N_BATCH,N_SAMPLED)
    samples_mass = tf.reduce_sum(samples_pred,axis=-1,keepdims=True) # (N_BATCH,1)

    with tf.variable_scope('grad_input_embed'):
      grad_input_embed = tf.matmul(samples_pred,samples_embed) # (N_BATCH,N_EMBED)
      grad_input_embed = grad_input_embed - samples_mass * labels_embed 
      grad_input_embed = tf.IndexedSlices(grad_input_embed, inputs, grad_input_embed_shape)

    with tf.variable_scope('grad_target_embed'):
      grad_target_embed1 = tf.matmul(samples_pred,inputs_embed,transpose_a=True) # (N_SAMPLED,N_EMBED)
      grad_target_embed2 = -samples_mass * inputs_embed # (N_BATCH,N_EMBED)
      grad_target_embed = tf.concat([grad_target_embed1,grad_target_embed2],0) # (N_BATCH+N_SAMPLED,N_EMBED)
      grad_target_embed = tf.IndexedSlices(grad_target_embed, tf.concat([samples,labels],0), grad_target_embed_shape)

\end{lstlisting}

\section{Performance Benchmarks}

We compare performance of the forward and backward pass for our and default tensorflow implementation. To this end we generated random data of size matching a typical SkipGram problem where the sampled loss is often used. The parameters are summarized in \Cref{tab:1}
\begin{table}[ht!]
    \centering
    \begin{tabular}{|c|c|c|c|}
    \hline
        classes & samples & embeded size & batch\\
    \hline
        100,000 & 100 & 300 & 256 \\
    \hline
    \end{tabular}
    \caption{Setup for our benchmark.}
    \label{tab:1}
\end{table}

The improvement is about 2 times for forward and backward pass, as illustrated on the graph below
\begin{figure}[th!]
\includegraphics[scale=0.35]{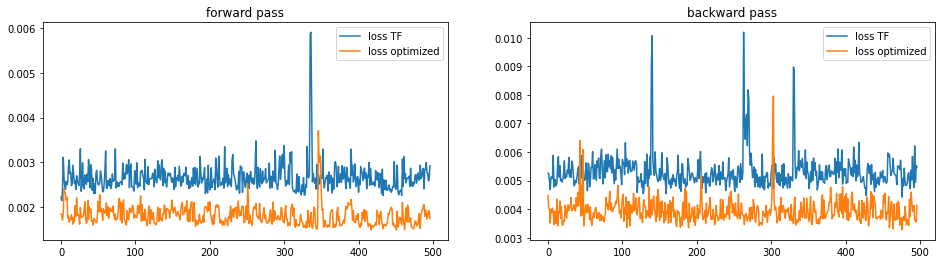}
\end{figure}

The code is included in the repo. The testing has been done in Google Colab.

\bibliographystyle{amsalpha}
\bibliography{citations}

\newcommand{\etalchar}[1]{$^{#1}$}
\providecommand{\bysame}{\leavevmode\hbox to3em{\hrulefill}\thinspace}
\providecommand{\MR}{\relax\ifhmode\unskip\space\fi MR }
\providecommand{\MRhref}[2]{%
  \href{http://www.ams.org/mathscinet-getitem?mr=#1}{#2}
}
\providecommand{\href}[2]{#2}
\begin{thebibliography}{AAB{\etalchar{+}}15}

\bibitem[AAB{\etalchar{+}}15]{tensorflow2015-whitepaper}
Mart\'{\i}n Abadi, Ashish Agarwal, Paul Barham, Eugene Brevdo, Zhifeng Chen,
  Craig Citro, Greg~S. Corrado, Andy Davis, Jeffrey Dean, Matthieu Devin,
  Sanjay Ghemawat, Ian Goodfellow, Andrew Harp, Geoffrey Irving, Michael Isard,
  Yangqing Jia, Rafal Jozefowicz, Lukasz Kaiser, Manjunath Kudlur, Josh
  Levenberg, Dandelion Man\'{e}, Rajat Monga, Sherry Moore, Derek Murray, Chris
  Olah, Mike Schuster, Jonathon Shlens, Benoit Steiner, Ilya Sutskever, Kunal
  Talwar, Paul Tucker, Vincent Vanhoucke, Vijay Vasudevan, Fernanda Vi\'{e}gas,
  Oriol Vinyals, Pete Warden, Martin Wattenberg, Martin Wicke, Yuan Yu, and
  Xiaoqiang Zheng, \emph{{TensorFlow}: Large-scale machine learning on
  heterogeneous systems}, 2015, Software available from tensorflow.org.

\bibitem[JCMB14]{jean2014using}
S{\'e}bastien Jean, Kyunghyun Cho, Roland Memisevic, and Yoshua Bengio,
  \emph{On using very large target vocabulary for neural machine translation},
  arXiv preprint arXiv:1412.2007 (2014).

\bibitem[Ten20]{tf_softmax}
Tensorflow,
  \url{https://www.tensorflow.org/api_docs/python/tf/nn/sampled_softmax_loss},
  2020.

\end{thebibliography}

\end{document}